%% file: main.tex
\documentclass[10pt,twocolumn,letterpaper]{article}

 \usepackage{cvpr}              
\usepackage{multicol}
\usepackage{booktabs}
\usepackage{multirow}
\usepackage{graphicx}
\usepackage[table]{xcolor}
\usepackage{bm}

\input{preamble}
\definecolor{cvprblue}{rgb}{0.21,0.49,0.74}
\usepackage[pagebackref,breaklinks,colorlinks,allcolors=cvprblue]{hyperref}

\def\confName{CVPR}
\def\confYear{2027}

\title{Beyond Reconstruction: Probing Visual Semantics in fMRI through Editing}

\author{
Zifu Zhang$^{1}$,
Xun Zhang$^{2}$,
Wei Xue$^{1}$,
Yike Guo$^{1}$,
Alessandro Bozzon$^{2}$,
Pan Wang$^{2\dagger}$ \\[6pt]
$^{1}$The Hong Kong University of Science and Technology \\
$^{2}$Delft University of Technology \\
}

\begin{document}
\maketitle
\input{fig/teaser}
\input{sec/0_abstract}    
\input{sec/1_intro}
\input{sec/2_semantic}

\input{sec/3_generalization}

\input{sec/4_discussion}
{
    \small
    \bibliographystyle{ieeenat_fullname}
    \bibliography{main}
}

\input{sec/X_suppl}

\end{document}

%% file: preamble.tex
\usepackage{cuted} 

\usepackage{currfile} 

\usepackage{caption} 

%% file: fig/teaser.tex
\preCutedStrip{\vskip -2em\relax}
\begin{strip}
\centering
\includegraphics[width=.93\linewidth]{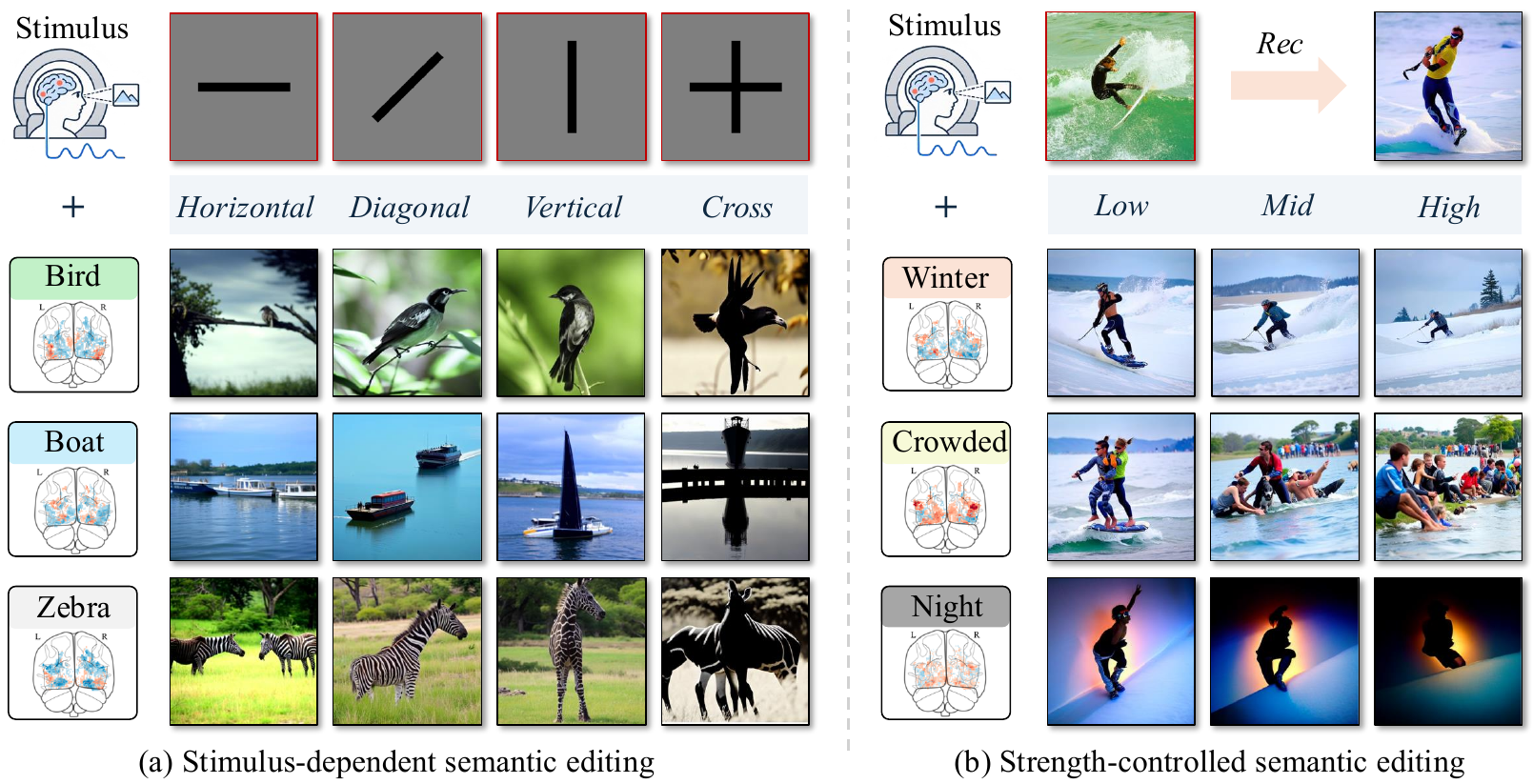}\hfill
\captionsetup{hypcap=false}
\vspace{-0.5em}
\captionof{figure}{Editing human fMRI representations can induce disentangled changes in reconstructed images, suggesting that these representations contain structured visual semantics.  (a) Edits interact with the structure of four geometric stimuli, producing stimulus-dependent outcomes. (b) Edits vary with editing strength for a fixed stimulus.}
\captionsetup{hypcap=true}
\label{fig:fig1}
\end{strip}

%% file: sec/0_abstract.tex
\begin{abstract}
Recent generative models have enabled high-fidelity reconstruction of perceived images from fMRI. However, high fidelity alone does not reveal whether fMRI signals encode visual information in a disentangled, spatially structured form, or whether the structure is largely supplied by the strong generalization of the generative prior. We investigate this question through semantic editing, in which concept-associated semantic fMRI, constructed from paired stimuli and fMRI responses, is applied directly to existing neural decoders without training any editing-specific module. Across four representative decoding frameworks with diverse encoders and generative backbones, the same semantic directions induce consistent semantic changes while largely preserving the pose and structure of the original reconstruction, providing evidence that fMRI representations carry semantic information that can be separately accessed and composed. We further find that editing only the image-conditioning branch achieves a better balance between semantic manipulation and content preservation than editing all conditioning branches. Building on this, we introduce a unified CLIP latent-space editing approach that remains competitive while removing the dependence on subject-specific voxel layouts. This enables the directions to transfer across subjects without additional training and from perception to mental imagery, where they induce target-related changes and improve the semantic alignment of imagery reconstructions. We hope semantic editing offers a complementary perspective for visual brain decoding, shifting the focus from how faithfully an image is reconstructed to how its visual factors are represented in brain signals and can be separately accessed, controlled, and composed.
\end{abstract}

%% file: sec/1_intro.tex
\vspace{-1.5em}
\section{Introduction}
\label{sec:intro}

Decoding visual experience from brain activity is a long-standing goal at the intersection of neuroscience and computer vision \cite{takagi2023high, liu2025survey}. Recent advances in large-scale generative models, in particular diffusion models \cite{ho2020denoising, rombach2022high}, have substantially advanced functional magnetic resonance imaging (fMRI) visual decoding, enabling reconstructions of perceived images with remarkable perceptual fidelity \cite{scotti2024mindeye2,scotti2023mindeye}. Most existing approaches map fMRI signals into pretrained semantic embedding spaces \cite{kneeland2026mirage, ozcelik2023braindiffuser}, such as CLIP image or text embeddings \cite{radford2021clip}, and use these representations to condition a frozen generator like unCLIP \cite{ramesh2022hierarchical} or versatile diffusion \cite{xu2023versatile} for image reconstruction. Consequently, progress has largely centered on designing better alignment between fMRI and one or more semantic spaces, while the generative model itself is treated as a fixed black box. This choice is well motivated by the broad generalization capabilities of generative models trained on large-scale image datasets \cite{deng2009imagenet} and the comparatively limited availability of fMRI data \cite{allen2022NSD}, which are costly to acquire.

Notably, recent work such as PRISM \cite{huang2026seeing} shows that conditioning on one-dimensional textual descriptions alone is sufficient to drive a pre-trained generator to produce high-fidelity reconstructions of two-dimensional images. This reveals that the spatial structure of the reconstruction may be supplied by the generative prior rather than by the brain signal. High perceptual fidelity therefore leaves a fundamental question unanswered: \textit{what exactly does an fMRI signal encode about the image, and can that information be accessed in a structured and factor-wise manner?} Existing evaluations cannot answer this question because holistic similarity does not distinguish between a model that recovers individual visual factors and one that merely retrieves a frequently co-occurring scene composition. For instance, when a method reconstructs ``a dog on snow'' high overall similarity does not tell us whether the identity of the dog, its pose, and the background are each recovered, or whether the model has mainly captured a common scene prior. Additionally, pixel-level metrics such as PSNR \cite{wang2004image}, which are standard in low-level vision \cite{li2025hierarchical}, are markedly lower for brain decoding, further suggesting that the fine-grained structure of the stimulus is only partially characterized. 

Semantic editing provides an operational test of whether visual factors encoded in fMRI signals can be selectively accessed and controlled. By applying concept-specific perturbations to fMRI signals or their derived representations, we can examine whether decoded images exhibit the intended semantic changes while preserving unrelated content. Consistent, selective changes would provide evidence for separately accessible semantic information in these representations, extending the analysis beyond overall reconstruction fidelity. This perspective is conceptually related to semantic editing in generative models, where manipulating latent representations helps reveal how visual attributes are organized and supports model interpretability \cite{shen2020interfacegan,harkonen2020ganspace,zhang2025improving, ju2024pnp}. For neural decoding, however, the critical requirement is that editing directions be derived from measured fMRI responses or their representations, rather than supplied solely through external conditioning at generation time.

In this paper, we study visual brain decoding from the perspective of fMRI editing and systematically investigate whether fMRI representations contain separable visual semantics. Using the publicly available fMRI dataset \cite{allen2022NSD, kneeland2025nsdimagery}, we construct semantic fMRI signals corresponding to different visual factors like bird, boat and zebra in Fig.~\ref{fig:fig1} via paired or concept-aggregated stimuli, and apply them directly to existing well-known decoders \cite{scotti2023mindeye, scotti2024mindeye2, ozcelik2023braindiffuser, kneeland2026mirage} without any training of editing-specific modules. Across multiple representative decoding frameworks, we observe consistent editing behavior, suggesting that the separability stems from the neural data rather than from a particular architecture. Building on this, we find that editing only the image branch yields a better balance between semantic manipulation and content preservation than direct editing in the original fMRI space. We then introduce a CLIP-space editing approach that maintains competitive editing performance while enabling direct reuse of semantic directions within a shared conditioning space. Finally, we examine generalization and find that the proposed editing transfers zero-shot across subjects and across tasks, from perception to mental imagery \cite{bergmann2024cortical}, indicating that the edited semantics are not subject-specific or task-specific artifacts. These results shift the question in brain decoding from how much of the image is recovered to whether its components can be separately accessed, controlled, and composed, and may suggest a complementary paradigm for future decoding research. Our main contributions are as follows:
\begin{itemize}
    \item We conduct a systematic study of semantic editing in fMRI-derived representations across multiple visual decoding frameworks, examining the consistency of concept-associated interventions.
    \item Building on our finding that image-branch-only editing better balances semantic control and content preservation, we propose a CLIP-space editing approach that achieves competitive performance while enabling semantic direction reuse within a shared conditioning space.
    \item We show that fMRI editing generalizes zero-shot across subjects and from perception to imagery, supporting the generality of the discovered semantic structure.
\end{itemize}

\section{Related Works}
\label{sec:relatedworks}

\subsection{Visual Neural Decoding}

Decoding visual experience from brain activity has been explored with a range of non-invasive modalities \cite{gao2026cinebrain}. EEG \cite{li2024visual} and MEG \cite{benchetrit2024brain} offer millisecond-level temporal resolution and low cost, and recent work has shown that they can support image retrieval and coarse reconstruction. However, their limited spatial resolution and low signal-to-noise ratio restrict the fine-grained visual information they can resolve \cite{li2024visual}. In contrast, fMRI measures whole-brain responses at millimeter-scale spatial resolution, and the release of large-scale datasets such as the NSD \cite{allen2022NSD} has made it the dominant modality for studying how visual content is represented and recovered. 

Recent fMRI decoding methods increasingly use pretrained generative models for image reconstruction. Brain-Diffuser \cite{ozcelik2023braindiffuser} combines low-level reconstruction with CLIP-based semantic guidance, while MindEye \cite{scotti2023mindeye} aligns fMRI with CLIP embeddings through contrastive learning and a diffusion prior. MindEye2 \cite{scotti2024mindeye2} and MindBridge \cite{wang2024mindbridge} improve cross-subject decoding through shared representations, and NeuroFlow \cite{mai2026neuroflow} unifies visual encoding and decoding through cross-modal flow matching. These methods advance reconstruction and representation alignment, but selective manipulation of individual visual factors remains comparatively underexplored.

\subsection{Brain-Guided Semantic Editing}

In generative vision, controllability has progressed from conditional generation \cite{zhang2023adding, mou2024t2i} to text-driven editing of real images \cite{hertz2022prompt, brooks2023instructpix2pix}. Beyond their practical utility, such studies also reveal how generative models organize visual attributes in their latent spaces, which motivates asking whether analogous structure can be found in brain signals. Existing efforts that connect brain signals with image editing fall into three groups. The first group transfers existing editing mechanisms to the neural modality through training. For example, BIE \cite{davis2022brain} uses EEG-derived supervision to learn GAN editing directions, while LoongX \cite{zhou2026neural} maps multimodal neurophysiological signals to editing conditions. These methods establish brain signals as an alternative input, but they learn a mapping from signals to editing conditions, and therefore do not reveal whether editable semantics are intrinsic to the neural data. The second group, represented by MindArt \cite{chen2024mind}, achieves stylized fMRI reconstruction, yet the editing content comes from external text prompts rather than from semantic information in the fMRI signal itself. The third group like BrainAlgebra \cite{ferrante2025evidence} directly manipulates fMRI responses, which constructs concept-associated perturbations from measured fMRI patterns and demonstrates compositional effects. However, it is built on a single decoding framework and covers only 12 predefined concepts, thus providing initial evidence rather than a systematic analysis. To the best of our knowledge, no prior work has systematically examined whether fMRI representations themselves contain separable visual semantics, independent of any specific decoding framework and without training editing-specific mappings.

%% file: sec/2_semantic.tex
\section{Semantic Editability Analysis}

We systematically investigate semantic editability in fMRI representations through three complementary analyses. First, we extract concept-associated directions from fMRI responses to examine whether individual visual factors can be selectively manipulated. Next, we apply these directions across multiple decoding frameworks to assess whether their effects are consistent across decoders. Finally, motivated by the limited semantic controllability of direct editing in the original fMRI space, we find that editing only the image branch yields a better balance between semantic manipulation and content preservation. Building on this finding, we introduce a CLIP latent-space editing approach, which maintains competitive performance while unifying the representation dimensionality across subjects, thereby facilitating cross-subject and cross-task studies.

\subsection{Semantic Extraction}
First, we extract semantic directions from paired images and fMRI responses in the Natural Scenes Dataset (NSD) \cite{allen2022NSD}. For each subject, the training set is denoted as
$\mathcal{D}_{T}=\{(\bm{I}_i,\bm{x}_i)\}_{i=1}^{N}$, which consists of $N$ pairs of a visual stimulus $\bm{I}_i$ and its voxel-wise standardized fMRI response $\bm{x}_i\in\mathcal{X}$. Both the standardization statistics and the semantic directions are estimated exclusively from this training set.

\begin{figure}[t]
    \centering
    \includegraphics[width=1\linewidth]{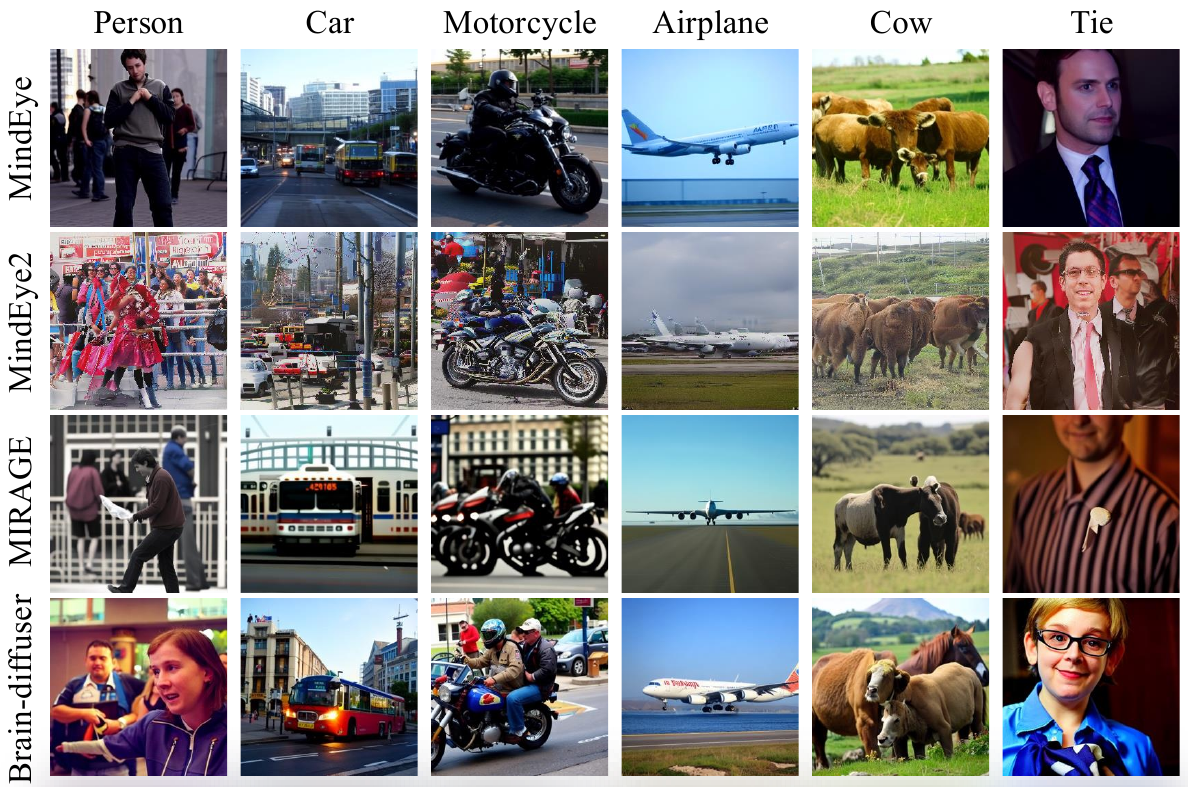}
    \vspace{-1em}
    \caption{Qualitative results of semantic editing fMRI $\bar{\bm{x}}_c$ across four neural decoding frameworks.}
    \vspace{-1em}
    \label{fig:semanticvector}
\end{figure}

For each target concept $c$ in a predefined semantic set $\mathcal{C}$, we identify the training stimuli most relevant to $c$ using a frozen CLIP model. We denote the normalized image and text features as $\phi(\bm{I}_i)$ and $\psi(c)$, respectively. To measure how strongly a stimulus expresses concept $c$ relative to the other concepts, we compute the target concept score $s^{c}_{i}$ as a softmax over the concept set $\mathcal{C}$, and select the $K$ highest-scoring in $\mathcal{D}_{T}$:
\begin{equation}
    s^{c}_{i} =\frac{\exp\!\left(\phi(\bm{I}_i)^\top\psi(c)\right)}{\sum_{c'\in\mathcal{C}}\exp\!\left(\phi(\bm{I}_i)^\top\psi(c')\right)},\quad
    \mathcal{S}_{c} =\operatorname{TopK}\!\left(s^{c}_{i}\right),
\end{equation}
where $\mathcal{S}_{c}$ denotes the index set of the top $K$ samples with the highest relative scores for concept $c$. Normalizing over $\mathcal{C}$ rather than using the raw similarity favors stimuli in which concept $c$ is distinctive, instead of stimuli that are generically similar to all concepts. Therefore, the fMRI responses of the selected samples are then averaged to obtain a concept-associated activation pattern $\bm{x}_c$:
\begin{equation}
\label{eq:x_c}
    \bm{x}_c = \frac{1}{K}\sum_{i\in\mathcal{S}_c}\bm{x}_i.
\end{equation}
 
To suppress weakly responsive voxels, we sparsify the concept-associated pattern by retaining only the voxels whose absolute activations exceed the $\lambda$-th percentile:
\begin{equation}
\label{eq:barxc}
    \bar{\bm{x}}_c =
    \bm{x}_c \odot \mathbb{I}\!\left(
    |\bm{x}_c| > Q_{\lambda}(|\bm{x}_c|)
    \right),
\end{equation}
where $\odot$ denotes element-wise multiplication, $\mathbb{I}(\cdot)$ is an indicator function applied across voxels, and $Q_{\lambda}(\cdot)$ denotes the empirical $\lambda$-th percentile of the absolute activations over all voxels. Voxels below the threshold are set to zero, while the retained voxels keep their original signs and magnitudes. The resulting $\bar{\bm{x}}_c$, visualized in Fig. \ref{fig:semanticvector}, serves as the final semantic editing fMRI for concept $c$. This sparsification keeps only the most concept-responsive voxels, and thereby reduces interference with non-target content during editing. We further implement the ROI analysis in Appendix~A, which supports the anatomical plausibility of the resulting semantic directions.

Given an original fMRI response $\bm{x}_0$, we apply the concept-associated perturbation with editing strength $\alpha$, and rescale the result so that it preserves the original $\ell_2$ norm:
\begin{equation}
\tilde{\bm{x}}^c_0 =
\frac{||\bm{x}_0||_2}
{||\bm{x}_0+\alpha\bar{\bm{x}}_c||_2}
\left(\bm{x}_0+\alpha\bar{\bm{x}}_c\right),
\end{equation}
where $\alpha$ controls the magnitude of the edit. The norm rescaling keeps $\tilde{\bm{x}}^c_0$ at the same scale as $\bm{x}_0$, so that the edited input stays within the range expected by the decoder and the observed changes can be attributed to the direction of the perturbation rather than to a change in signal energy.

The edited fMRI response $\tilde{\bm{x}}_0$ is passed through the fMRI encoder $E$ and the generator $G$ to produce the edited image:
\begin{equation}
\tilde{\bm{I}}^c_0=G\left(E(\tilde{\bm{x}}^c_0)\right),
\end{equation}
where $E$ maps the fMRI response into the conditioning space of the generator, and $G$ denotes the frozen generative model. All parameters of $E$ and $G$ remain fixed throughout editing, so that no editing-specific module is trained and any change in $\tilde{\bm{I}}^c_0$ arises solely from the perturbation applied to the input. By varying $\alpha$, we examine whether the extracted direction induces the intended semantic change in the decoded image, and to what extent the remaining visual attributes are preserved.

To evaluate the extracted semantic directions, we use fMRI responses from the NSD-Imagery~\cite{kneeland2025nsdimagery} vision dataset as base representations $\bm{x}_0$, in particular the oriented bars and crosses in Set A, whose simple and well-defined geometry facilitates examining how added semantic content interacts with stimulus-related structure. As shown in Fig.~\ref{fig:fig1}(a), applying the same noun direction, drawn from the semantic set $\mathcal{C}$ of 80 COCO category labels, to different base representations yields concept-related images with distinct spatial configurations, suggesting that the added category information combines with structural information retained in the base representation. Fig.~\ref{fig:fig1}(b) further shows that varying the strength of adjective directions, extracted from constructed antonym pairs, modulates seasonal context, scene density, and lighting, although stronger edits may also alter other attributes. Together, these observations provide qualitative evidence that fMRI representations contain semantic information that can be selectively accessed.

\subsection{Cross-Framework Consistency}

\begin{figure}[b]
    \centering
    \vspace{-1em}
    \includegraphics[width=1\linewidth]{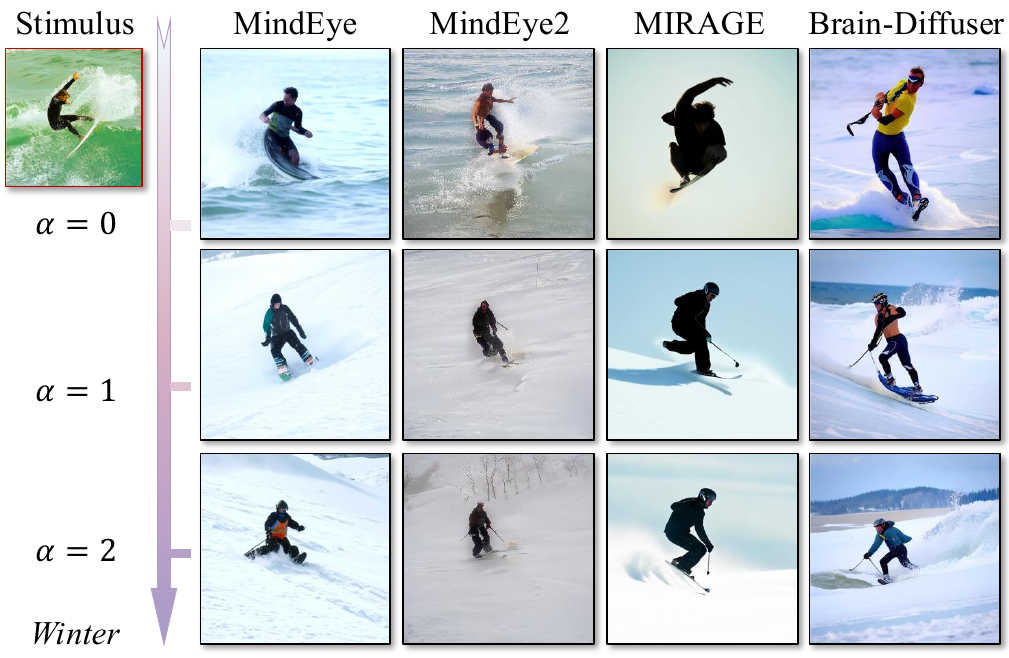}
    \vspace{-1em}
    \caption{Semantic fMRI editing across four neural decoding frameworks. For the same stimulus, fMRI vectors are edited with increasing strength $\alpha=\{0,1,2\}$ along the winter semantic direction. Results at $\alpha=0$ show unedited reconstructions.}
    \label{fig:allframeworks}
\end{figure}

\begin{figure*}[ht]
    \centering
    \includegraphics[width=0.95\linewidth]{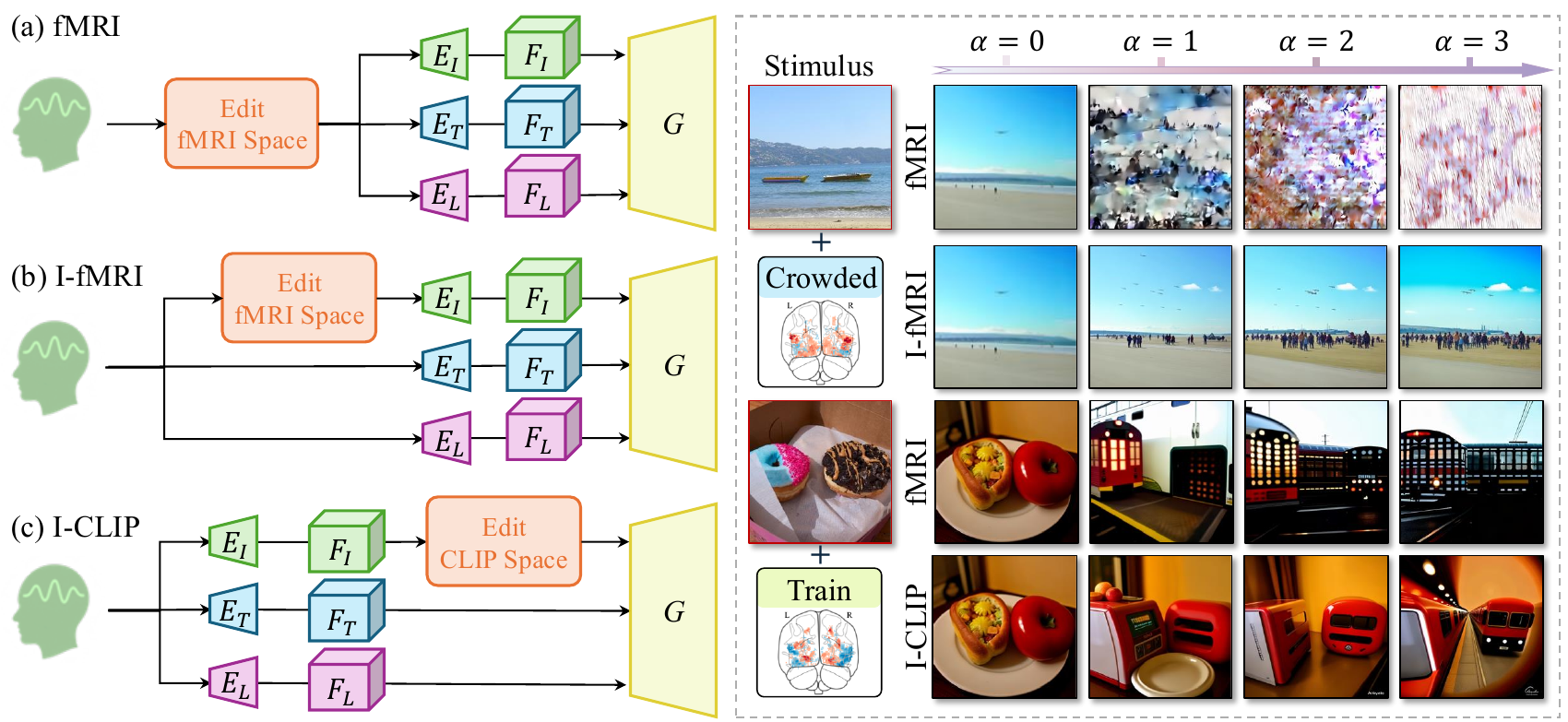}
    \vspace{-0.5em}
    \caption{Comparison of editing configurations using MIRAGE~\cite{kneeland2026mirage}. {(a)} edits all conditioning branches in voxel space, {(b)} edits only the image branch in voxel space, and {(c)} edits the CLIP image representation. Compared with all-branch editing, which tends to degrade image quality or replace the source scene, image-only editing better preserves the source layout while integrating the target semantics.}
    \label{fig:space}
    \vspace{-1em}
\end{figure*}

Since semantic extraction and editing are performed entirely in fMRI space, the resulting perturbations can be applied before decoding without modifying the downstream architecture. We therefore evaluate whether the same semantic directions produce consistent effects across four decoding frameworks that differ in both encoding architecture and generative backbone, with all decoding parameters held fixed. MindEye~\cite{scotti2023mindeye} combines an MLP backbone with a diffusion prior and generates images with Versatile Diffusion~\cite{xu2023versatile}, while MindEye2~\cite{scotti2024mindeye2} introduces shared-subject modeling and uses SDXL~\cite{podell2024sdxl}. MIRAGE~\cite{kneeland2026mirage} applies ridge regression to predict multimodal conditions for Stable Cascade~\cite{pernias2023wuerstchen}. Brain-Diffuser~\cite{ozcelik2023braindiffuser} also relies on ridge regression, predicting VDVAE~\cite{child2020very} latents and CLIP features that guide reconstruction with Versatile Diffusion. This diversity allows us to examine whether semantic editability extends beyond a particular encoding or generation pipeline. Fig.~\ref{fig:semanticvector} shows that the extracted concept fMRI $\bar{\bm{x}}_c$, when fed directly into a decoder without any extra input, yield recognizable category-specific images in all four frameworks, despite their different encoding architectures and generative backbones. This indicates that the category information carried by $\bar{\bm{x}}_c$ is not tied to a particular framework.

\definecolor{BrainEditBlue}{HTML}{E7F2FC}
\definecolor{BrainEditGreen}{HTML}{E8F4E9}
\definecolor{BrainEditGray}{HTML}{F8F8F8}

\begin{table*}[t]
\centering
\vspace{-0.5em}
\caption{Quantitative comparison of three editing strategies at matched editing strengths. {fMRI} denotes the original BrainAlgebra editing applied to all conditioning branches, {I-fMRI} applies fMRI-space editing only to the image-conditioning branch, and {I-CLIP} directly edits the image condition in the CLIP latent space. CLIP-I (similarity to the unedited reconstruction) measures content preservation, CLIP-T (alignment with the target concept) measures editing effectiveness, and PixCorr and SSIM (similarity to the original stimulus) measure stimulus consistency. Higher is better for all metrics, and the best result within each group of three is shown in bold.}
\vspace{-0.5em}
\label{tab:editing}
\resizebox{0.97\linewidth}{!}{%
\begin{tabular}{@{}cc*{4}{ccc}@{}}
\toprule
\multirow{2}{*}{Framework} & \multirow{2}{*}{Strength}
  & \multicolumn{3}{c!{\hspace{4pt}}}{\cellcolor{BrainEditBlue}CLIP-I$\uparrow$}
  & \multicolumn{3}{c!{\hspace{4pt}}}{\cellcolor{BrainEditGreen}CLIP-T$\uparrow$}
  & \multicolumn{3}{c!{\hspace{4pt}}}{\cellcolor{BrainEditGray}PixCorr$\uparrow$}
  & \multicolumn{3}{c}{\cellcolor{BrainEditGray}SSIM$\uparrow$} \\
\cmidrule(lr){3-5}\cmidrule(lr){6-8}
\cmidrule(lr){9-11}\cmidrule(lr){12-14}
 & & fMRI & I-fMRI & I-CLIP
   & fMRI & I-fMRI & I-CLIP
   & fMRI & I-fMRI & I-CLIP
   & fMRI & I-fMRI & I-CLIP \\
\midrule
\multirow{2}{*}{MindEye} & low
 & 65.94\% & \textbf{67.71\%} & 64.27\%
 & 22.81\% & 22.81\% & \textbf{24.00\%}
 & 0.252 & 0.294 & \textbf{0.295}
 & 0.383 & 0.389 & \textbf{0.403} \\
 & high
 & 58.97\% & \textbf{60.33\%} & 57.91\%
 & 23.88\% & 23.81\% & \textbf{24.81\%}
 & 0.182 & \textbf{0.280} & 0.268
 & 0.359 & 0.376 & \textbf{0.392} \\
\addlinespace[2pt]
\multirow{2}{*}{MindEye2} & low
 & 68.15\% & 68.41\% & \textbf{76.20\%}
 & 23.48\% & 23.52\% & \textbf{23.86\%}
 & 0.165 & \textbf{0.180} & 0.176
 & 0.389 & 0.398 & \textbf{0.420} \\
 & high
 & 59.70\% & 60.27\% & \textbf{61.05\%}
 & 24.90\% & \textbf{24.99\%} & 24.92\%
 & 0.119 & \textbf{0.153} & 0.104
 & 0.369 & \textbf{0.389} & 0.373 \\
\addlinespace[2pt]
\multirow{2}{*}{MIRAGE} & low
 & 75.38\% & \textbf{78.60\%} & 78.35\%
 & {22.22\%} & 22.15\% & \textbf{22.33}\%
 & 0.261 & \textbf{0.266} & \textbf{0.266}
 & \textbf{0.441} & 0.437 & {0.437} \\
 & high
 & 71.05\% & \textbf{73.42\%} & 73.17\%
 & {22.93\%} & 23.15\% & \textbf{23.21}\%
 & 0.248 & \textbf{0.262} & 0.260
 & 0.430 & \textbf{0.431} & 0.429 \\
\addlinespace[2pt]
\multirow{2}{*}{Brain-Diffuser} & low
 & 69.45\% & 70.86\% & \textbf{70.92\%}
 & {20.25\%} & \textbf{20.35}\% & 20.26\%
 & 0.134 & 0.139 & \textbf{0.140}
 & 0.371 & 0.373 & \textbf{0.376} \\
 & high
 & 62.46\% & \textbf{62.48\%} & 62.40\%
 & \textbf{22.01\%} & 21.61\% & 21.57\%
 & 0.116 & 0.138 & \textbf{0.139}
 & 0.351 & 0.361 & \textbf{0.370} \\
\bottomrule
\end{tabular}%
}
\vspace{-1em}
\end{table*}

We then examine whether such concept fMRI can modify a specific factor of an existing reconstruction, as shown in Fig.~\ref{fig:allframeworks}. Starting from the same stimulus, we increase $\alpha$ along the \emph{winter} direction. Although the unedited reconstructions ($\alpha=0$) differ markedly across frameworks in pose, scene, and style, all four progressively shift their background toward snowy scenes as $\alpha$ grows, while largely preserving a person engaged in a board sport. As $\alpha$ increases, the activity shifts naturally from water sports to skiing or snowboarding, indicating that the winter direction combines coherently with the existing semantic content rather than overwriting it. Meanwhile, the edited images largely retain the pose of the unedited reconstructions, suggesting that the edit alters the scene context while leaving subject-related structure mostly intact. Comparisons of edits for additional concepts are provided in Appendix~B.

\subsection{Editing Strategies Comparison}

Directly manipulating fMRI representations enables semantic editing, but its effects become unstable as the editing strength increases (Fig.~\ref{fig:space}). We attribute this to two factors. First, fMRI-space perturbations applied to all conditioning branches also alter the text and low-level conditions, introducing changes unrelated to the target semantics. Second, perturbations must propagate through the encoding pipeline, which may turn additive changes into irregular displacements in the conditioning space. We therefore compare three strategies, namely {fMRI}, which edits all conditioning branches in voxel space, {I-fMRI}, which edits only the image branch in voxel space, and {I-CLIP}, which edits the CLIP image representation directly. The last is motivated by generative image editing in the CLIP space \cite{patashnik2021styleclip, kim2022diffusionclip}.

For I-CLIP, based on the extracted semantic fMRI $\bar{\bm{x}}_c$ in Eq.~(\ref{eq:barxc}), we construct a latent semantic direction by encoding it with the frozen mapping $E_I$ from fMRI responses to the CLIP image-conditioning space, and subtracting the encoding of a zero baseline:
\begin{equation}
\label{eq:z_c}
\bar{\bm{z}}_c = E_I(\bar{\bm{x}}_c)-E_I(\bm{0}).
\end{equation}
Subtracting $E_I(\bm{0})$ removes the component that $E$ assigns to an uninformative input, so that $\bar{\bm{z}}_c$ represents the concept as an offset relative to the baseline. The perturbation is then applied to $\bm{z}_0 = E_I(\bm{x}_0)$ with norm preservation:
\begin{equation}
\tilde{\bm{z}}_0^{c}=\frac{\|\bm{z}_0\|_F}
{\|\bm{z}_0+\alpha\bar{\bm{z}}_c\|_F}
\left(\bm{z}_0+\alpha\bar{\bm{z}}_c\right),
\end{equation}
where the Frobenius norm is taken over the complete conditioning tensor, including all tokens and feature dimensions. The edited image is obtained as $\tilde{\bm{I}}_0^{c}=G(\tilde{\bm{z}}_0^{c})$. Only the CLIP image condition is modified, while all other conditioning inputs and model parameters remain fixed. I-fMRI follows the same procedure but applies the perturbation to $\bm{x}_0$ in voxel space and passes it only to the image branch.

Fig.~\ref{fig:space} shows that restricting edits to the image branch avoids the failure modes of all-branch editing. For \emph{crowded}, increasing $\alpha$ in all-branch fMRI editing disrupts the beach structure and introduces substantial artifacts, whereas I-fMRI preserves the beach layout while increasing the number of people. For \emph{train}, all-branch editing largely replaces the source scene with a railway scene, whereas I-CLIP integrates train-related appearance into the source composition. Overall, image-only editing better preserves the source layout while integrating the target semantics. Additional quantitative results are supplied in Appendix~C.

Table~\ref{tab:editing} quantifies this comparison at matched editing strengths. Editing only the image branch markedly improves content preservation, with I-fMRI achieving higher CLIP-I than all-branch fMRI editing in nearly all settings, without sacrificing target alignment. Building on this, CLIP space editing yields competitive metrics relative to I-fMRI, and attains the best CLIP-T for MindEye and MIRAGE and the best CLIP-I for MindEye2. Its advantages over I-fMRI are therefore moderate and framework-dependent. Additionally, the CLIP space editing no longer depends on the voxel layout of the target subject, and the resulting direction $\bar{\bm{z}}_c$ can be cached and reused across base samples, subjects, and tasks. This makes it a more convenient basis for the semantic transferability study in Section~\ref{sec:transfer}, and we adopt it as the default editing strategy.

%% file: sec/3_generalization.tex
\section{Semantic Transferability}
\label{sec:transfer}

Having shown that semantic directions extracted from fMRI support editing across decoding frameworks, we next ask whether these directions are tied to the subject and task from which they were derived. If the edited semantics reflect general properties of visual representation rather than idiosyncrasies of a particular subject or paradigm, the same directions should remain effective beyond their source. Since editing is performed in a CLIP conditioning space shared across subjects, we examine this question from two perspectives. The first is transfer across subjects, without specific training of any editing component. The second is transfer across tasks, from perception to mental imagery, which further allows us to test whether editing can serve as a practical means of enhancing reconstruction in imagery decoding.

\begin{figure}[t]
    \centering
    \includegraphics[width=0.95\linewidth]{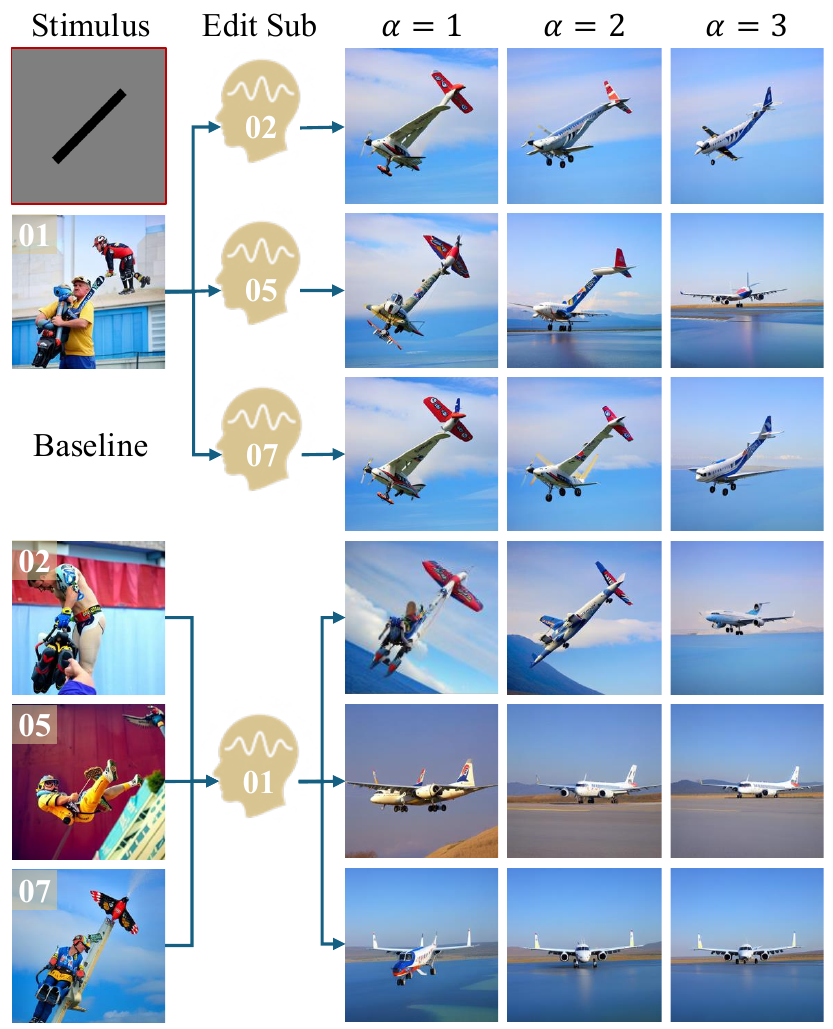}
    \caption{Cross-subject transfer of the \emph{airplane} direction based on Brain-Diffuser. In the top part, latent directions extracted from subjects 02, 05, and 07 are applied to subject 01, whereas in the bottom part, the direction extracted from subject 01 is applied to the other three subjects. Additional results are shown in Appendix~D.}
    \label{fig:crosssub2}
    \vspace{-1em}
\end{figure}

\subsection{Cross-Subject Transfer}

Cross-subject generalization is a central challenge in brain decoding, since fMRI responses differ across individuals in voxel dimensionality and anatomical correspondence, which precludes direct transfer of semantic directions in the original fMRI space. Within a given decoding framework, however, subject-specific encoders map these heterogeneous inputs into a common CLIP representation space. This shared space provides a common coordinate system in which semantic directions may be transferred without additional alignment training.

We denote the source and target subjects by $s$ and $t$, respectively. Following the latent-space editing formulation in Eq. (\ref{eq:z_c}), the semantic direction is extracted with the source subject’s encoder $E^s_I$, and the base representation is obtained with the target subject’s encoder $E^t_I$:
\begin{equation}
\bar{\bm{z}}_c^{s}
=
E^s_I(\bar{\bm{x}}_c^{s})-E^s_I(\bm{0}),
\quad
\bm{z}_0^{t}=E^t_I(\bm{x}_0^{t}).
\end{equation}
The source direction is then applied to the target representation with norm preservation:
\begin{equation}
\tilde{\bm{z}}_0^{c}
=
\frac{\|\bm{z}_0^{t}\|_F}
{\|\bm{z}_0^{t}+\alpha\bar{\bm{z}}_c^{s}\|_F}
\left(
\bm{z}_0^{t}+\alpha\bar{\bm{z}}_c^{s}
\right).
\end{equation}
The edited representation is decoded by the target subject’s generation pipeline, with all other conditions held fixed. Since the source direction is applied without refitting or rescaling, and only pre-existing subject-specific decoders are used, this constitutes zero-shot transfer of the editing direction.

Fig.~\ref{fig:crosssub2} illustrates bidirectional transfer of the \emph{airplane} direction between subject 01 and subjects 02, 05, and 07. Although the baseline reconstructions of the same diagonal-bar stimulus differ substantially across subjects, directions extracted from other subjects introduce airplane-related structures and outdoor context into subject 01’s reconstructions. Conversely, the direction extracted from subject 01 induces analogous changes in the reconstructions of the other three subjects. The transferred effects vary in appearance and strength but share a common semantic tendency. These results suggest that the extracted directions capture semantic information that remains usable across subjects, despite differences in original representations.

\subsection{Cross-Task Transfer}

Mental imagery constitutes a substantial distribution shift from visually evoked brain activity \cite{roy2025transformation, horikawa2017generic, anderson2025neural}. In the absence of external visual input, imagery-related fMRI responses exhibit lower signal-to-noise ratios and less precise spatial information, which makes reconstruction particularly challenging. Accordingly, strong performance on perceptual decoding does not necessarily carry over to imagery decoding \cite{kneeland2025nsdimagery}. We therefore examine whether semantic directions extracted from perception data remain effective when applied to imagery representations, and whether they can further be used to enhance the semantic content of imagery reconstructions.

\begin{figure}[t]
    \centering
    \includegraphics[width=1\linewidth]{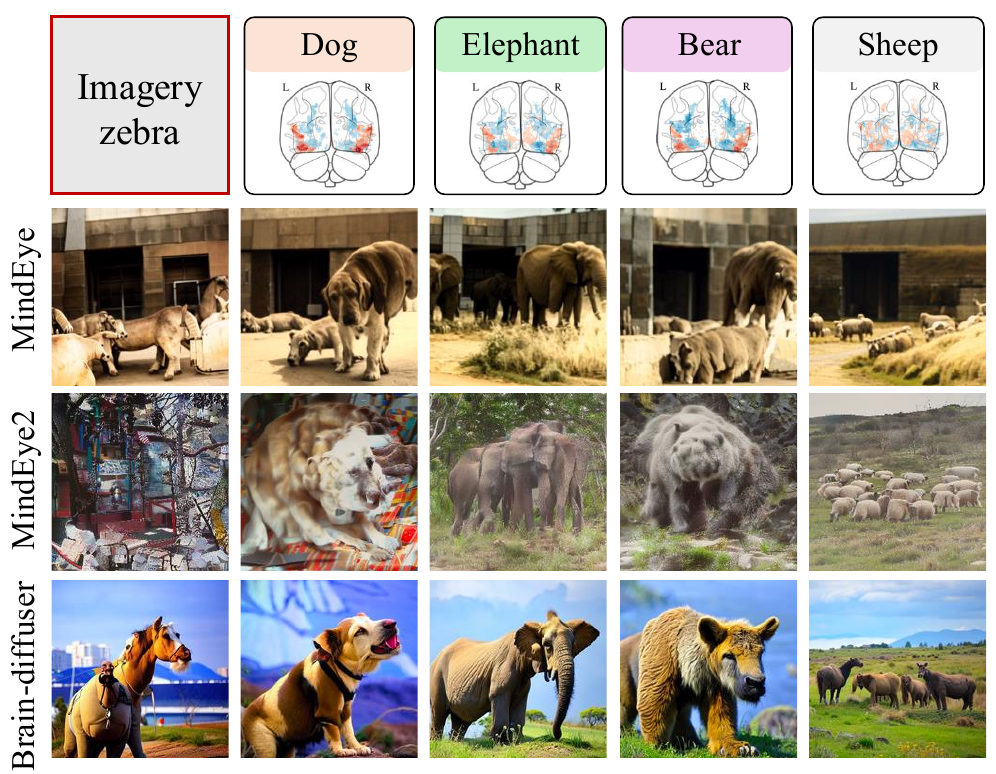}
    \vspace{-1.5em}
    \caption{Semantic editing of imagery responses to the \emph{zebra} cue in NSD-Imagery \cite{kneeland2025nsdimagery}. The first column shows the unedited reconstructions, while the remaining columns show edits obtained with perception-derived animal concepts. Target-related semantic changes remain observable across three decoding frameworks.}
    \label{fig:imagery}
    \vspace{-1em}
\end{figure}

\begin{table}[b]
\centering
\vspace{-1em}
\caption{Comparison of fMRI input and edit direction in cue-guided
imagery enhancement with MIRAGE based on NSD-Imagery Set C. Absolute scores measure alignment
with the target cue, and margins measure selectivity relative to
alternative cues.}
\vspace{-0.5em}
\label{tab:mirage_s01_imagery}
\resizebox{\linewidth}{!}{
\begin{tabular}{@{}llcccc@{}}
\toprule
\multirow{2}{*}{Input} & \multirow{2}{*}{Edit}
 & \multicolumn{2}{c}{CLIP-T} & \multicolumn{2}{c}{Brain Corr} \\
\cmidrule(lr){3-4}\cmidrule(lr){5-6}
& & Score $\uparrow$ & Margin $\uparrow$
                    &     Score $\uparrow$ & Margin $\uparrow$ \\
\midrule
Original       & None   & 21.09\% & 0.55\% & 3.92\% & 4.71\% \\
Train mean & fMRI   & 24.23\% & 4.61\% & 4.42\% & 5.18\% \\
Shuffled   & fMRI   & 24.22\% & 4.00\% & 2.78\% & 3.27\% \\
Original       & Visual & 24.97\% & 5.29\% & 3.45\% & 4.08\% \\
Original       & fMRI   & \textbf{25.44\%} & \textbf{5.48\%}  & \textbf{4.85\%} & \textbf{5.73\%} \\
\bottomrule
\end{tabular}}
\end{table}

We evaluate this transfer on NSD-Imagery by applying perception-derived directions to the CLIP representations of imagery responses, without retraining the decoders or re-estimating semantic vectors from imagery data. As shown in Fig.~\ref{fig:imagery}, the unedited reconstructions for the \emph{zebra} cue differ markedly across frameworks and fail to recover the intended category. Applying the \emph{dog}, \emph{elephant}, \emph{bear}, and \emph{sheep} directions nevertheless induces target-related changes in most cases, such as the body shape of elephants and the flock structure of sheep, while parts of the source scene, including the outdoor setting, are retained. This indicates that perception-derived directions remain effective under the imagery task despite the shift in input distribution. However, edit quality depends on the base reconstruction. When the MindEye2 baseline is largely unstructured, for instance, the edit yields only coarse shapes and colors. Semantic editing can thus inject category-level information into imagery reconstructions, but cannot fully compensate for structure missing from the base reconstruction.

Beyond changing the target concept, we investigate whether editing can reinforce the semantics of the original imagery stimulus. To test this, we compare the effect of the input and the edit direction in cue-guided imagery enhancement with MIRAGE, as shown in Table~\ref{tab:mirage_s01_imagery}. Replacing the true imagery response with a training-mean fMRI or a shuffled fMRI, while keeping the same cue-specific edit, raises CLIP-T to a similar level. This confirms that the edit direction alone injects substantial cue-related semantics, and that CLIP-T score by itself cannot separate the contribution of the input. However, the input matters for selectivity and for agreement with brain activity. Editing the original imagery fMRI achieves the highest CLIP-T score and margin. The difference is clearer under Brain Corr metric \cite{st2022brain}. A shuffled input carries the wrong subject-trial information, and the edit cannot compensate for it in brain-aligned terms. This indicates that the edit does not simply override the input but composes with the information it carries. Finally, the fMRI-derived direction outperforms a visual direction built from the same selected images, suggesting that the direction constructed from measured brain responses is better aligned with the brain-decoding pathway than one derived from the images alone. These results suggest that the enhancement arises from the combination of a cue direction and the true imagery representation, rather than from the direction alone. See Appendix~E for further results.

%% file: sec/4_discussion.tex
\section{Ablation Study}
\subsection{Semantic fMRI Construction}

\begin{table}[b]
\centering
\vspace{-1em}
\caption{Ablation of the number of averaged samples $K$ and the sparsification percentile $\lambda$ in semantic vector construction.}
\vspace{-0.5em}
\label{tab:mindeye-vision-ablation-alpha2}
\resizebox{1\linewidth}{!}{%
\begin{tabular}{lccccc}
\toprule
Setting & CLIP-I$\uparrow$ & CLIP-T$\uparrow$ & PixCorr$\uparrow$ & SSIM$\uparrow$ & AlexNet(2)$\uparrow$ \\
\midrule
$K=1$   & \textbf{59.19\%} & 24.50\% & 0.264 & 0.376 & 64.51\% \\
$K=50$  & 58.03\% & 24.73\% & 0.266 & 0.385 & \textbf{66.26\%} \\
$K=100$ & 57.91\% & \textbf{24.81\%} & \textbf{0.268} & \textbf{0.392} & 65.68\% \\
$K=150$ & 58.67\% & 23.99\% & 0.266 & 0.384 & 65.73\% \\
\midrule
$\lambda=0$  & 58.06\% & 24.44\% & 0.265 & 0.385 & \textbf{67.25\%} \\
$\lambda=50$ & \textbf{58.41\%} & 24.43\% & 0.267 & 0.385 & 66.67\% \\
$\lambda=75$ & 57.91\% & \textbf{24.81\%} & \textbf{0.268} & \textbf{0.392} & 65.68\% \\
$\lambda=95$ & 57.41\% & 23.52\% & 0.266 & 0.390 & 65.38\% \\
\bottomrule
\end{tabular}}
\vspace{-1em}
\end{table}

Two hyperparameters govern the construction of the semantic fMRI: the number of averaged samples $K$ in Eq.~(\ref{eq:x_c}), and the sparsification percentile $\lambda$ in Eq.~(\ref{eq:barxc}), which determines how many low-magnitude voxels are discarded. We ablate each parameter while fixing the other at its default ($\lambda=75$ when varying $K$, and $K=100$ when varying $\lambda$), and report the results in Table~\ref{tab:mindeye-vision-ablation-alpha2}. The results are averaged over 13 concepts using MindEye at $\alpha=2$. Varying $K$ reveals a trade-off between noise and specificity. A single sample retains sample-specific noise and yields lower CLIP-T and SSIM, whereas averaging too many samples dilutes concept-specific information and reduces CLIP-T to 23.99\%. $K=100$ achieves the best CLIP-T, PixCorr, and SSIM. A similar pattern holds for $\lambda$. Without sparsification, concept-irrelevant voxels interfere with the edit, while overly aggressive filtering discards concept-relevant responses and sharply lowers CLIP-T to 23.52\%. $\lambda=75$ yields the best CLIP-T, PixCorr, and SSIM. We therefore adopt $K=100$ and $\lambda=75$, as they provide the balance between semantic alignment and stimulus preservation.

\subsection{Selector Choice}

To examine whether the extracted directions depend on the selector, we replace CLIP with SigLIP \cite{zhai2023sigmoid} when selecting the stimuli used to construct $\bar{\bm{x}}_c$, and keep all other components fixed. As shown in Table~\ref{tab:siglip-selector}, the two selectors yield highly similar results at both editing strengths. SigLIP reduces CLIP-T by only 0.24 and 0.90 percentage points, which is consistent with a slight advantage of CLIP when the evaluation metric belongs to the same model family, while CLIP-I, SSIM, and the angular change of the image condition remain nearly unchanged. These results indicate that the effectiveness of the directions does not hinge on the specific model, and that the semantic information they carry is not an artifact of the selector.
\begin{table}[ht]
\centering
\vspace{-0.5em}
\caption{Selector ablation for MindEye. Angle denotes the angular change of the image condition.}
\vspace{-0.75em}
\label{tab:siglip-selector}
\setlength{\tabcolsep}{4pt}
\resizebox{\columnwidth}{!}{%
\begin{tabular}{@{}ccccccc@{}}
\toprule
Edit & Selector & CLIP-T$\uparrow$ & CLIP-I$\uparrow$ & SSIM$\uparrow$ & AlexNet(2)$\uparrow$ & Angle \\
\midrule
\multirow{2}{*}{$\alpha=1$}
  & CLIP   & {24.00\%} & {64.27\%} & 0.403 & 72.38\% & 41.43$^\circ$ \\
  & SigLIP & 23.76\% & 63.84\% & 0.416 & 72.96\% & 41.64$^\circ$ \\
\addlinespace[3pt]
\multirow{2}{*}{$\alpha=2$}
  & CLIP   & 24.81\% & 57.91\% & 0.392 & 65.68\% & 63.06$^\circ$ \\
  & SigLIP & 23.91\% & 57.67\% & 0.399 & 64.10\% & 62.72$^\circ$ \\
\bottomrule
\end{tabular}%
}
\end{table}

\vspace{-1em}
\section{Conclusion}
In this paper, we studied visual brain decoding from the perspective of fMRI editing, asking whether fMRI representations contain separable visual semantics. By applying concept-associated directions constructed from paired stimuli and fMRI responses to fixed decoders, we observed consistent semantic edits across four decoding frameworks, providing evidence of factor-wise accessible information in the signal. Editing only the image-conditioning branch achieves a better balance between semantic manipulation and content preservation, and editing directly in the CLIP latent space remains competitive while removing the dependence on subject-specific voxel layouts. This allows the directions to transfer across subjects without additional training and from perception to imagery, where they enhance semantic alignment. We hope that semantic editing can serve as a complementary lens for visual brain decoding, shifting the focus from how faithfully an image is reconstructed to how its visual factors are represented in brain signals, and how they can be separately accessed, controlled, and composed. Such a view may inform future decoders that treat reconstruction as the composition of semantic factors rather than a single holistic mapping.


\section*{LLM Usage Statement}
We used ChatGPT to polish the manuscript's language and grammar, and Codex to assist with code development. Their use was limited to writing refinement and implementation assistance. All research ideas and methodological contributions were independently developed by the authors.

%% file: sec/X_suppl.tex
\clearpage
\setcounter{page}{1}
\maketitlesupplementary
\appendix

\section{Cortical ROI Analysis}
\label{app:roi_analysis}

\begin{figure*}[t]
    \centering
    \includegraphics[width=1\linewidth]{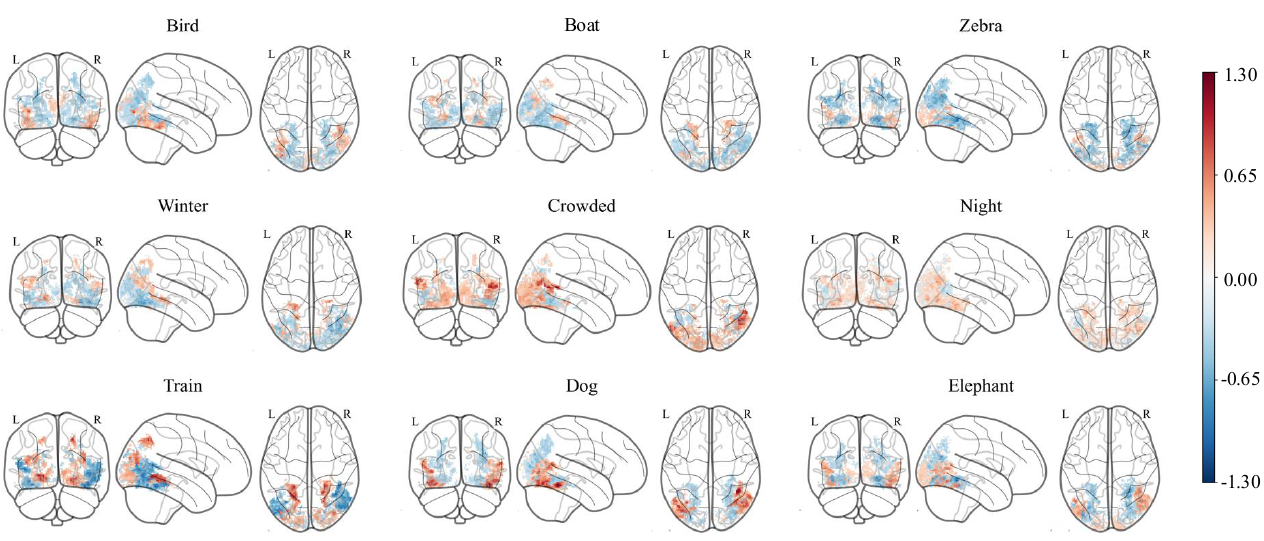}
    \caption{Cortical distribution of semantic directions. Each panel shows the signed weights of a concept direction $\bar{\bm{x}}_c$ projected onto the brain, with warm and cool colors indicating positive and negative weights.}
    \label{fig:brainfig}
\end{figure*}

We examine whether the extracted semantic directions spatially correspond to established functional subdivisions of visual cortex. This analysis characterizes the anatomical distribution of the editing weights, without assuming that individual concepts are localized to distinct regions. Each direction $\bar{\bm{x}}_c$ is mapped back to the native 1.8-mm functional volume using the original voxel indexing, over an input mask $V$ of 15,724 voxels for subject 01. Following the extraction procedure, each direction retains 3,931 nonzero entries after percentile thresholding on absolute weights. Regions of interest are taken from the official subject-specific NSD collections, comprising face-, body-, and place-selective ROIs as well as retinotopic and visual-stream parcellations, of which 22 regions intersect with $V$ and are included in the analysis. The visualization results are shown in Fig. \ref{fig:brainfig} and all measurements are computed in native space, without MNI resampling or additional smoothing.

For a region of interest (ROI) $R$, we denote by $R_V=R\cap V$ the subset of its voxels covered by the input mask. We first compute the mean signed weight of the direction within the ROI:
\begin{equation}
\mu_{c,R}
=
\frac{1}{|R_V|}
\sum_{v\in R_V} [\bar{\bm{x}}_c]_v .
\label{eq:roi_signed_mean}
\end{equation}
To characterize how strongly the weights concentrate in an ROI, while accounting for ROI size and for differences in the overall magnitude of each direction, we further define the absolute-weight enrichment as the ratio between the ROI-wise and mask-wide mean absolute weights:
\begin{equation}
\eta_{c,R}
=
\frac{
\displaystyle
\frac{1}{|R_V|}
\sum{v\in R_V}
\left|[\bar{\bm{x}}_c]_v\right|
}{
\displaystyle
\frac{1}{|V|}
\sum{v\in V}
\left|[\bar{\bm{x}}_c]_v\right|
} .
\label{eq:roi_enrichment}
\end{equation}
Both quantities are computed over all voxels in the corresponding set, including those zeroed out by thresholding. An enrichment $\eta_{c,R}>1$ indicates that the ROI carries a larger mean absolute weight than the mask-wide average. This measure is descriptive and does not imply statistical significance. Moreover, because it is based on absolute values, it does not reflect the polarity of the weights, so whether an ROI shows a positive preference for a category is assessed separately through the signed mean $\mu_{c,R}$.

\begin{figure}[t]
    \centering
    \includegraphics[width=\linewidth]{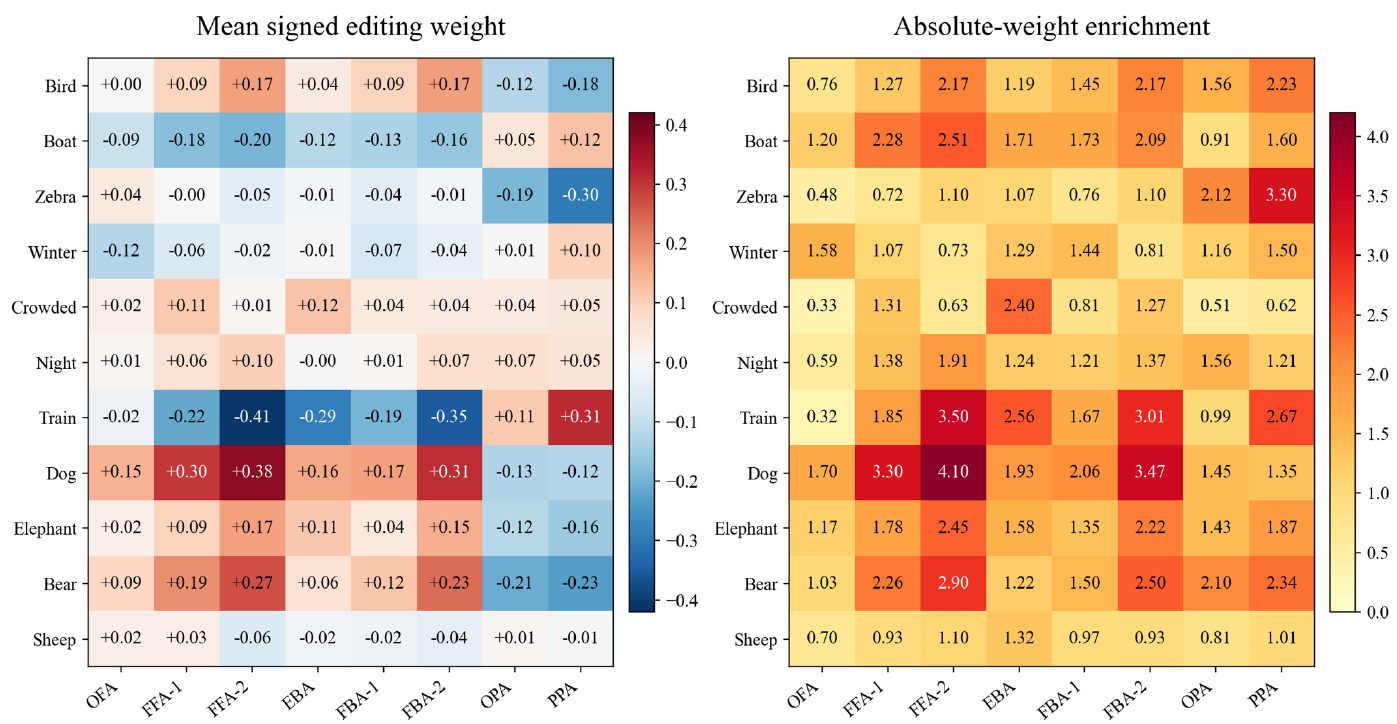}
    \caption{
    Native-space correspondence between semantic editing
    directions and functional visual ROIs in subject~01.
    Left: mean signed weights $\mu_{c,R}$.
    Right: absolute-weight enrichment $\eta_{c,R}$ relative
    to the complete nsdgeneral input mask.
    Measurements use the available portion of each ROI and
    include zero-valued entries after thresholding.
    These are descriptive editing-weight statistics,
    not activation significance maps.
    }
    \label{fig:roi_correspondence}
\end{figure}

Fig.~\ref{fig:roi_correspondence} shows that the directions are distributed over visual cortex, with partial correspondence to functional organization. The \emph{crowded} direction is enriched in the extrastriate body area (EBA; $\eta_{c,R}=2.40$, $\mu_{c,R}=0.118$), compatible with a contribution from body-related content. Several animal directions carry positive weights in face- and body-selective ventral regions. In FFA-2, \emph{dog}, \emph{elephant}, and \emph{bear} reach enrichments of $4.10$, $2.45$, and $2.90$ with positive mean weights of $0.376$, $0.171$, and $0.269$, and \emph{bird} shows about $2.17$-fold enrichment in both FFA-2 and FBA-2. These patterns indicate correspondence with regions defined by face and body localizers, rather than animal-specific selectivity. \emph{Train}, \emph{boat}, and \emph{winter} show positive mean weights in the parahippocampal place area (PPA), with enrichments of $2.67$, $1.60$, and $1.50$, which may reflect scene and spatial-context information in the images used to build the directions.

The correspondence is not uniform across concepts. \emph{Zebra} is strongly enriched in PPA ($\eta_{c,R}=3.30$) but has a negative mean weight ($\mu_{c,R}=-0.298$), \emph{sheep} lacks the clear face- and body-ROI pattern seen for other animals, and \emph{night} shows a mixed profile across scene- and face-selective regions. These results support the anatomical plausibility of some extracted directions.

\begin{figure}[!t]
    \centering
    \includegraphics[width=1\linewidth]{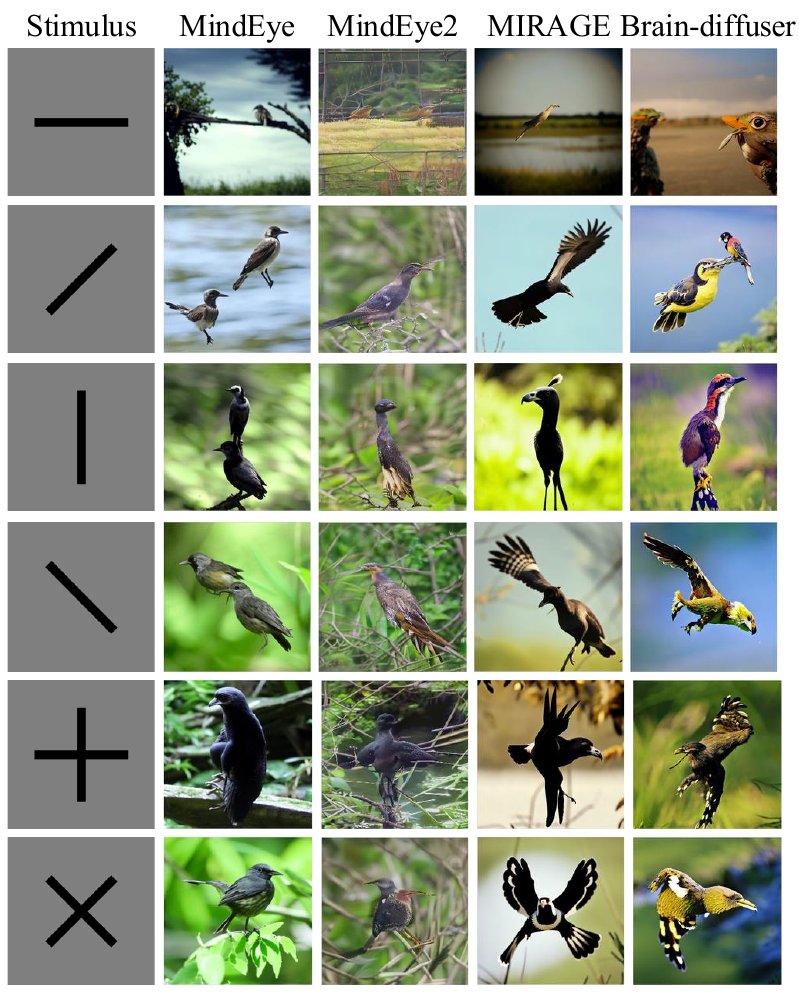}
    \caption{Bird-guided semantic editing across four brain decoders. Each row shows a geometric stimulus and the corresponding edited outputs, illustrating variation in bird appearance and pose across stimuli and decoders.}
    \label{fig:app-b1}
\end{figure}

\section{Cross-Framework Analysis}
\label{app:crossframework_analysis}

\subsection{Implementation Details}

Table~\ref{tab:frameworks} summarizes the semantic conditioning and image generators used in our experiments. MindEye and MindEye2 employ image embeddings aligned through a diffusion prior, whereas MIRAGE and Brain-Diffuser use both image and text embeddings decoded from fMRI. These frameworks also incorporate low-level information: MindEye initializes img2img sampling using an SD VAE-based reconstruction, while MIRAGE and Brain-Diffuser use VDVAE-based reconstructions for initialization. In our MindEye2 configuration, the low-level reconstruction is instead blended with the SDXL unCLIP output using weights of 0.25 and 0.75, respectively; caption-guided SDXL refinement
is not enabled.

\begin{table}[t]
\centering
\caption{Semantic conditioning and image generators used by the reconstruction frameworks in our experiments.}
\label{tab:frameworks}
\resizebox{\linewidth}{!}{%
\begin{tabular}{@{}lll@{}}
\toprule
Framework & Condition & Generator \\
\midrule
MindEye        & CLIP image, Low-level Feature        & Versatile Diffusion \\
MindEye2       & CLIP image, Low-level Feature    & SDXL unCLIP \\
MIRAGE         & CLIP image, CLIP text, Low-level Feature & Stable Cascade \\
Brain-Diffuser & CLIP image, CLIP text, Low-level Feature & Versatile Diffusion \\
\bottomrule
\end{tabular}}
\end{table}

\subsection{Additional Results}

\begin{figure}[t]
    \centering
    \includegraphics[width=1\linewidth]{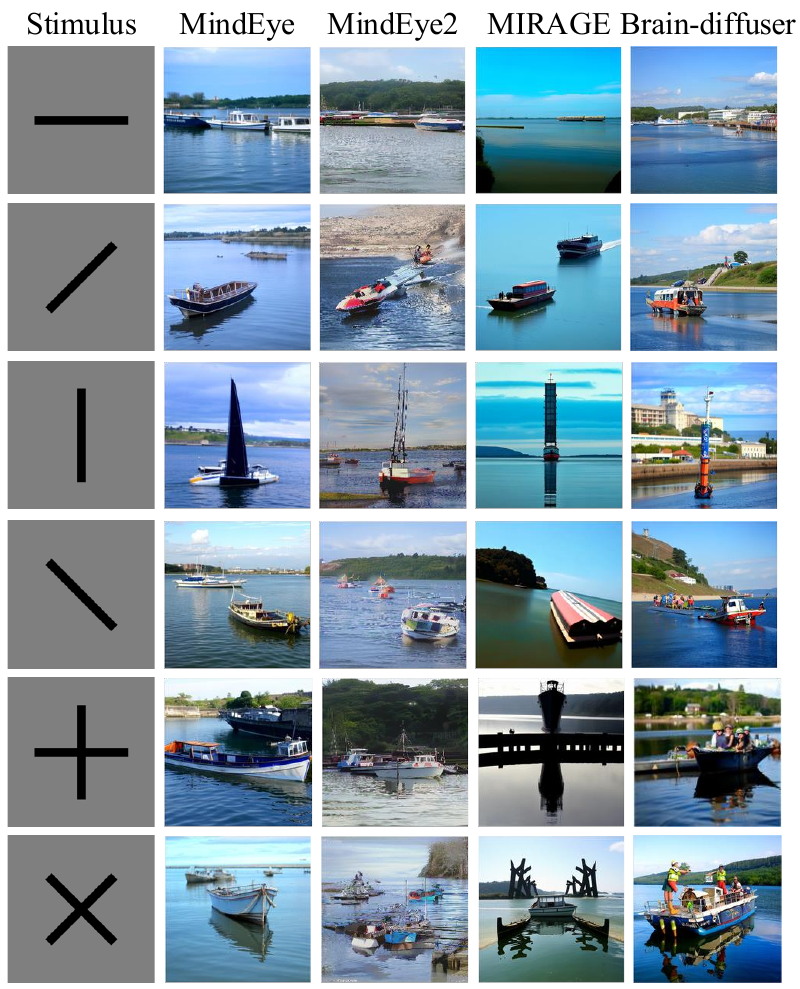}
    \caption{Boat-guided semantic editing across four brain decoders. Each row shows a geometric stimulus and the corresponding edited outputs, illustrating variation in boat appearance and pose across stimuli and decoders.}
    \label{fig:app-b2}
\end{figure}

\begin{figure}[ht]
    \centering
    \includegraphics[width=1\linewidth]{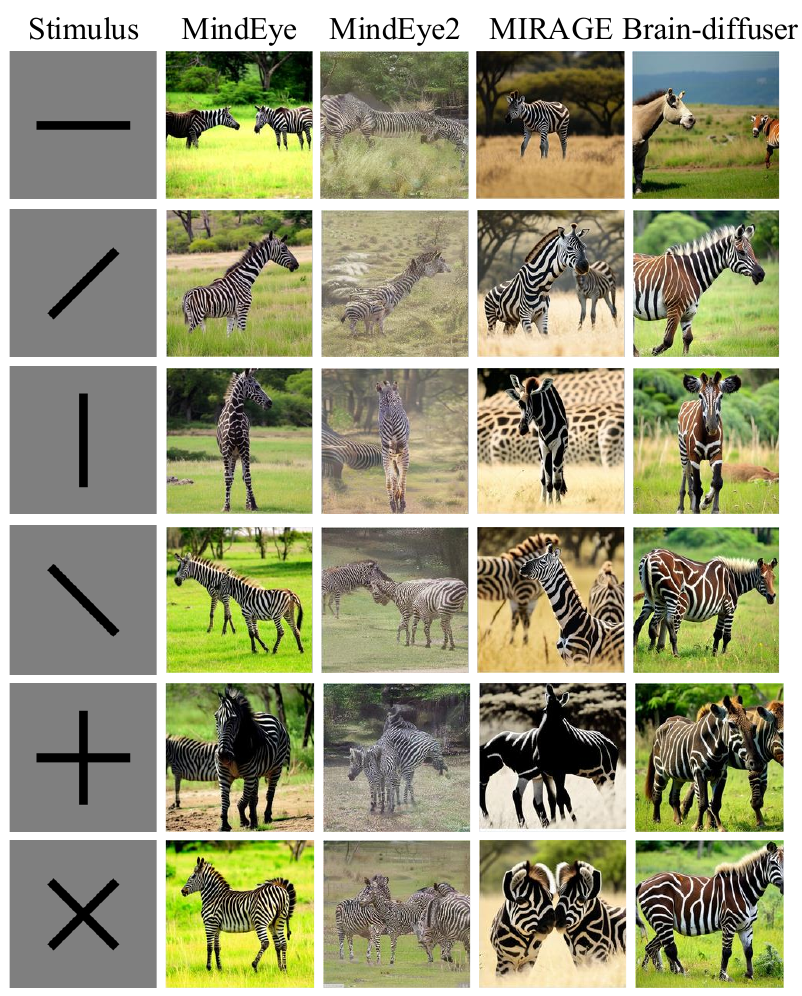}
    \caption{Zebra-guided semantic editing across four brain decoders. Each row shows a geometric stimulus and the corresponding edited outputs, illustrating variation in zebra appearance and pose across stimuli and decoders.}
    \label{fig:app-b3}
\end{figure}

Figures~\ref{fig:app-b1}--\ref{fig:app-b3} present additional qualitative results of semantic editing with \emph{bird}, \emph{boat}, and \emph{zebra} directions across four decoding frameworks, namely MindEye, MindEye2, MIRAGE, and Brain-Diffuser. In all frameworks, the edited outputs adopt the target category while retaining geometric cues from the original stimulus, such as object orientation, pose, silhouette, or spatial arrangement, rather than literal geometric symbols. The recurrence of this behavior across semantic targets and frameworks supports the feasibility of decoupling semantic content from stimulus geometry.

The correspondence is most apparent for simple line stimuli. Horizontal, vertical, and diagonal inputs tend to induce matching directional structure in the edited images, expressed through bird poses, boat configurations, or zebra body orientations. A shared geometric cue can thus be instantiated by different categories, and the same category can accommodate different stimulus geometries. Although appearance and rendering style vary across frameworks, the overall pattern is consistent.

More complex stimuli, such as plus and cross symbols, show differences in the clarity of this correspondence. For example, under bird guidance, the cross stimulus yields a clearly recognizable intersecting configuration in MIRAGE, whereas the Brain-Diffuser output conveys the bird semantics but corresponds less explicitly to the cross geometry. Such cases reflect variation in the degree of geometric preservation across frameworks rather than a lack of decoupling capability in any one of them. Overall, these examples suggest that all four frameworks support semantic--geometric decoupling to varying degrees, with fidelity depending on the stimulus, target category, and decoder.

\section{Editing Strategies Comparison}

\subsection{Comparison with BrainAlgebra}

Our work extends BrainAlgebra in three aspects. First, we introduce controlled geometric stimuli to examine how semantic concepts combine with spatial structure. By making orientation and shape explicit, these experiments assess semantic--geometric composition more directly than semantic transformations of natural scenes alone, and help distinguish structure-preserving editing from wholesale content replacement. Second, we broaden the evaluation across decoding frameworks, subjects, and both perception and imagery tasks, including explicit transfer of editing directions between subjects. Together with quantitative comparisons of semantic alignment and reconstruction fidelity, this characterizes the consistency and limitations of editing beyond individual qualitative examples. Third, we extend the original fMRI-space intervention by examining both the conditioning branch and the representation space in which editing is performed. Specifically, we introduce image-branch-only fMRI editing (I-fMRI) and norm-preserving editing of the final CLIP image condition (I-CLIP), which provide more selective control over how semantic modifications enter the reconstruction pipeline.

\begin{figure}[t]
    \centering
    \includegraphics[width=1\linewidth]{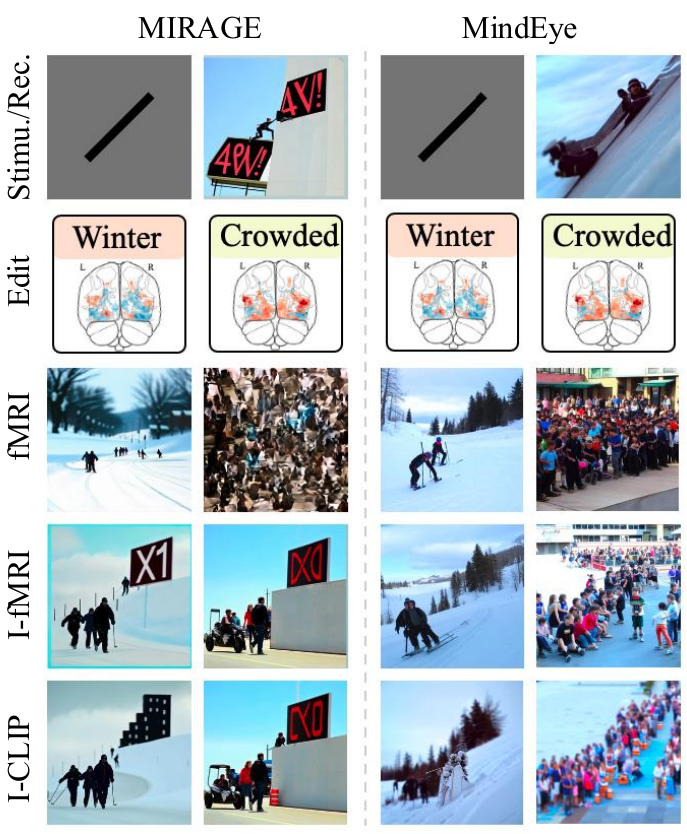}
    \caption{Comparison of editing strategies using MIRAGE and MindEye with the \emph{winter} and \emph{crowded} targets. We compare three editing strategies of all-branch fMRI editing (fMRI), image-branch-only fMRI editing (I-fMRI), and norm-preserving editing of the CLIP image condition (I-CLIP). Restricting edits to the image branch better preserves source content, while I-CLIP balances target semantics and scene layout.}
    \label{fig:app-c}
\end{figure}

\subsection{Additional Results}

Fig.~\ref{fig:app-c} provides additional comparisons under two conditioning configurations. In our implementations, MIRAGE combines CLIP image conditioning, CLIP text conditioning, and low-level image initialization, whereas MindEye uses CLIP image conditioning and low-level image initialization. The fMRI strategy propagates the modified brain representation through all supported conditioning branches, I-fMRI restricts the modification to the image branch, and I-CLIP directly edits the final image condition while preserving its norm and leaving the other branches unchanged. These strategies differ markedly in practice. Full fMRI editing can produce strong target semantics but substantially alters the original scene, as seen in the \emph{crowded} examples. Restricting the intervention to the image branch improves content preservation, with MIRAGE retaining the distinctive sign-and-wall setting and MindEye better maintaining the spatial organization of the scene. I-CLIP offers the most favorable qualitative balance between expressing the target concept and retaining source content and layout, which is particularly evident in the \emph{crowded} examples, where the original scene remains recognizable while additional people are introduced. These observations support selective, norm-preserving editing of the image condition, although fine details and sharpness still depend on the decoder and example.

\section{Additional Cross-Subject Results}

\begin{figure}[t]
    \centering
    \includegraphics[width=1\linewidth]{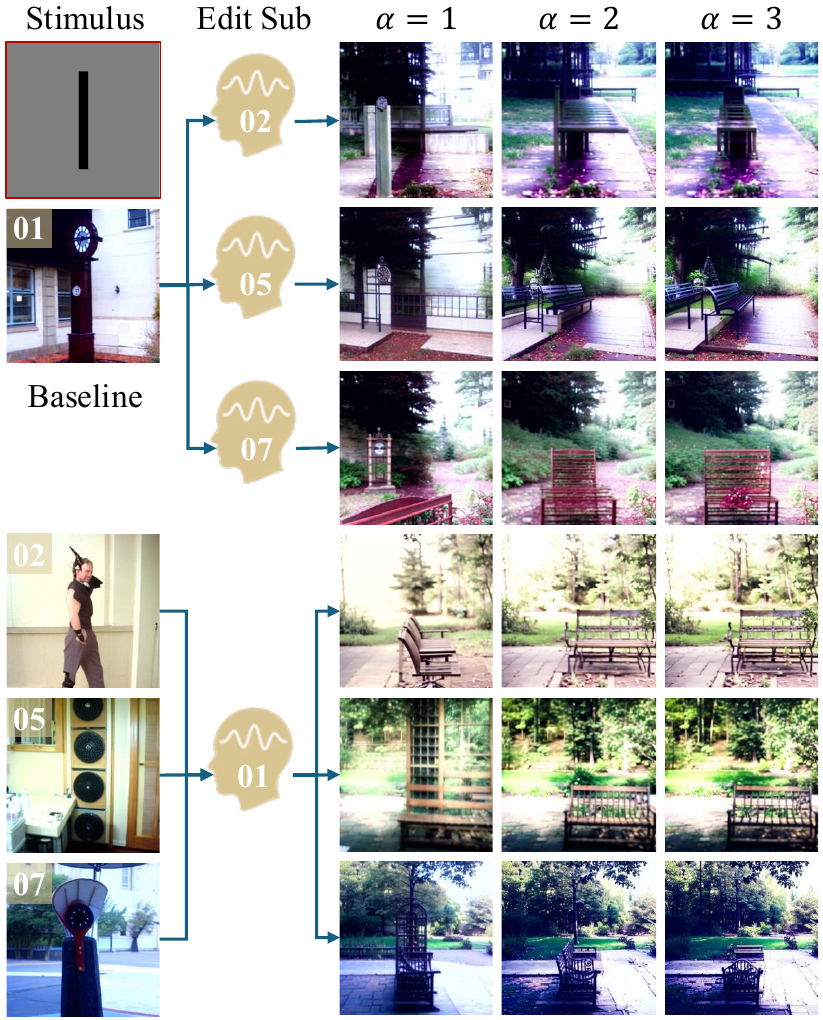}
    \caption{Cross-subject transfer of the \emph{bench} direction based on MindEye. In the top part, latent directions extracted from subjects 02, 05, and 07 are applied to subject 01, whereas in the bottom part, the direction extracted from subject 01 is applied to the other three subjects.}
    \label{fig:app-d}
\end{figure}

We provide additional cross-subject results using MindEye and the \textit{bench} editing direction in Fig.~\ref{fig:app-d}. We evaluate bidirectional transfer between subject 01 and subjects 02, 05, and 07 with the same vertical-bar stimulus. In the top part, directions derived from subjects 02, 05, and 07 are applied to subject 01, and in the bottom part, the direction derived from subject 01 is applied to the other three subjects. Each direction is computed with the source subject’s own mapping and applied in the target subject’s CLIP image-conditioning space, while the target decoder and all other conditioning branches remain unchanged. Editing strengths are $\alpha \in \{1,2,3\}$.

Despite substantial differences in the unedited reconstructions, the transferred directions introduce bench-related structures in all six source--target pairs, with more recognizable benches generally emerging at $\alpha=2$ and $\alpha=3$. Different source directions applied to subject 01 yield distinct bench appearances and surrounding scenes, and the subject-01 direction likewise produces different outcomes across target subjects. This suggests that the transferred direction carries a shared semantic tendency, while the resulting image is also shaped by the source direction and the target reconstruction pipeline. Some outputs retain a prominent vertical organization, but stronger edits can substantially alter the original content and background. These examples therefore support the feasibility of cross-subject direction transfer, without implying identical effects across subjects or consistently preserved stimulus geometry.

\section{Additional Imagery Results}

Beyond changing the target concept, we investigate whether semantic editing can strengthen the expression of the original imagery concept in decoded reconstructions. We extend this evaluation to four decoding frameworks, namely MindEye, MindEye2, MIRAGE, and Brain-Diffuser. For subject 01, we consider all six concepts in Set C, namely \emph{stripes}, \emph{zebra}, \emph{mammal}, \emph{yellow}, \emph{banana}, and \emph{fruit}. For each imagery response, we select the perception-derived direction corresponding to its original cue and apply norm-preserving editing of the CLIP image condition, while leaving the other conditioning branches unchanged. We compare unedited reconstructions ($\alpha=0$) with edited outputs at $\alpha=2$ and $\alpha=4$, which tests whether directions estimated from perception data can also enhance concept expression in imagery reconstructions across decoding architectures.

\begin{figure}[ht]
    \centering
    \includegraphics[width=1\linewidth]{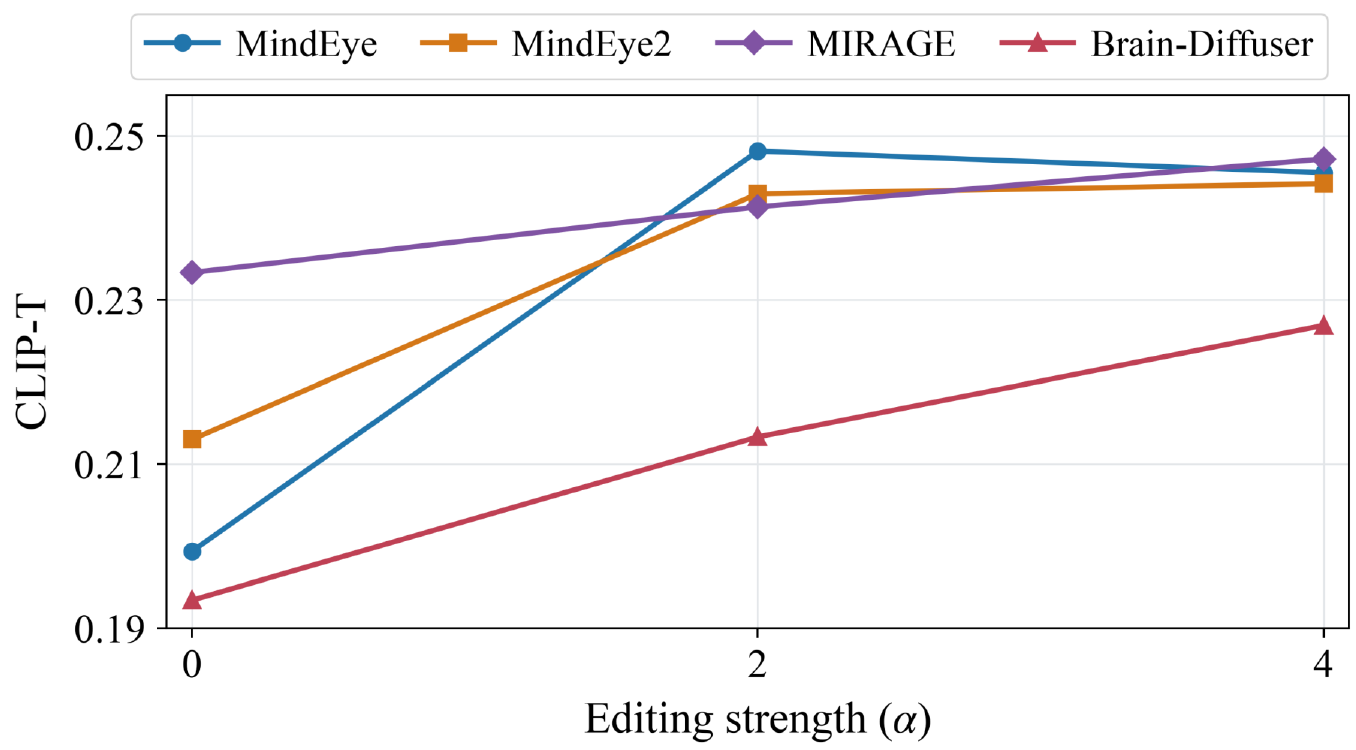}
    \caption{Semantic enhancement of imagery reconstructions. Each imagery response is edited along the direction corresponding to its original cue, and CLIP-T similarity to that cue is averaged over the Set C concepts. The editing strengths improve the mean alignment over the unedited baseline across all frameworks.}
    \label{fig:app-e1}
\end{figure}

Fig.~\ref{fig:app-e1} reports the CLIP-T similarity between each reconstruction and its original imagery cue, averaged equally over the six concepts. All four frameworks exceed their unedited baselines at both editing strengths. MindEye improves from 0.199 to 0.248 and 0.245 at $\alpha=2$ and $\alpha=4$, respectively, and MindEye2 from 0.213 to 0.243 and 0.244. MIRAGE, which has the highest baseline, increases from 0.233 to 0.241 and 0.247, and Brain-Diffuser from 0.193 to 0.213 and 0.227. These consistent gains indicate that the procedure is applicable beyond a single decoder. The response curves also show that editing strength interacts with the framework. Stronger editing continues to raise the mean score for several frameworks, whereas MindEye peaks at $\alpha=2$, so a larger strength does not necessarily bring further improvement.

\begin{figure}[t]
    \centering
    \includegraphics[width=1\linewidth]{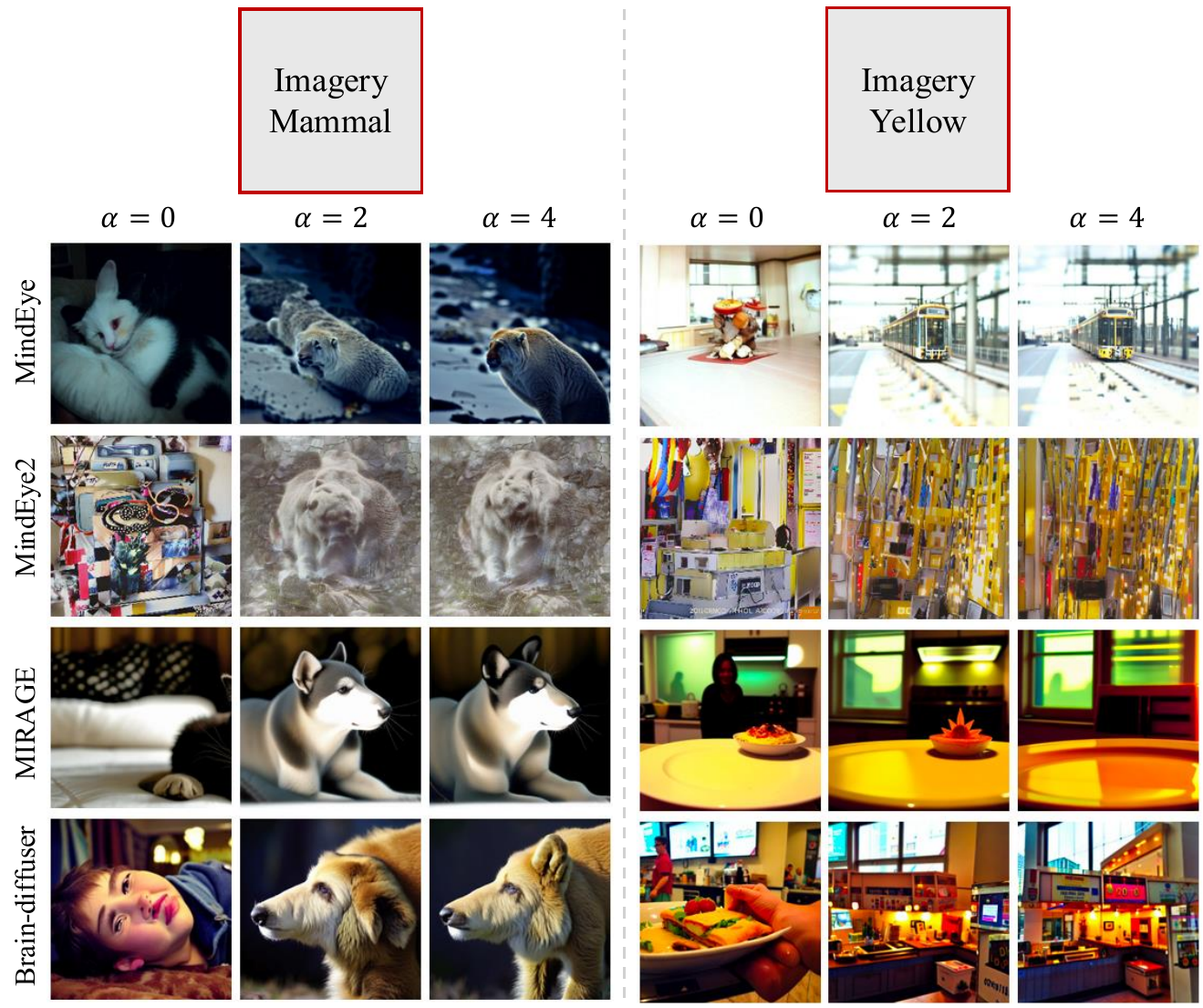}
    \caption{Qualitative examples of semantic enhancement for imagery reconstructions with the imagery cues \emph{mammal} and \emph{yellow}. Rows correspond to MindEye, MindEye2, MIRAGE, and Brain-Diffuser. Within each concept group, columns show the unedited reconstruction ($\alpha=0$) and the edited outputs at $\alpha=2$ and $\alpha=4$. All examples are from subject 01 and are included in the quantitative evaluation in Fig.~\ref{fig:app-e1}. Editing along the corresponding perception-derived direction yields more recognizable mammalian subjects or more prominent yellow content, with decoder-dependent changes in appearance and scene composition.}
    \label{fig:app-e2}
\end{figure}

Fig.~\ref{fig:app-e2} complements these results with examples for \emph{mammal} and \emph{yellow}. The unedited outputs differ substantially across frameworks in appearance and in how clearly they express the intended concept. For \emph{mammal}, editing yields recognizable animal subjects in all four frameworks, and some reconstructions that initially contain unrelated objects or human-centered scenes shift toward explicit non-human mammals. Where an animal is already present, editing can change its appearance, pose, or species rather than merely sharpening it. For \emph{yellow}, the effect appears as yellow objects, surfaces, or more prominent yellow coloration. A shared semantic objective is thus realized through different visual changes depending on the decoder and its initial reconstruction.

Overall, these results indicate that perception-derived directions can provide useful control over imagery reconstructions, improving their average alignment with the intended concepts across frameworks, particularly when the unedited reconstruction expresses the cue weakly or ambiguously. The aggregate improvement does not imply that every concept benefits equally, and stronger semantic expression can accompany changes in object identity or scene composition. Moreover, since the direction is selected using the known imagery cue, these results establish the effectiveness of cue-guided semantic enhancement, but not more faithful recovery of unknown mental content or of the precise scene imagined by the participant.